\pdfoutput=1 
\documentclass[10pt,letterpaper]{article}

\usepackage{cogsci}

\cogscifinalcopy 

\usepackage{graphicx}
\usepackage{amsmath}
\usepackage{pslatex}
\usepackage{apacite}
\usepackage{float} 

{\obeyspaces\gdef {\ }}
\global\newbox\codebox
\global\newbox\savedcodebox
\gdef\sverbatim{\bgroup\def\endsverbatim{\egroup\egroup\egroup\mbox{\box\codebo\
x}}\def\savecode{\egroup\egroup\egroup\global\setbox\savedcodebox\copy\codebox}\
\def\par{\egroup\vspace{-0.3em}\hbox\bgroup}\tt\obeylines\obeyspaces\global\set\
box\codebox\vbox\bgroup\hbox\bgroup}

\newcommand{\avmplus}[1]{{\setlength{\arraycolsep}{0.8mm}
                       \renewcommand{\arraystretch}{1.2}
                       \left[
                       \begin{array}{l}
                       \\[-2mm] #1 \\[-2mm] \\
                       \end{array}
                       \right]
                    }}
\newcommand{\att}[1]{{\mbox{\small {\bf #1}}}}
\newcommand{\attval}[2]{{\mbox{\small {\sc #1}}\ =\ {{#2}}}}

\newcommand{\attvaltyp}[2]{{\mbox{\small{\sc #1}}\ =\ {\myvaluebold{#2}}}}

\newcommand{\myvaluebold}[1]{{\mbox{\small {\bf #1}}}}

\newenvironment{enumerate*}%
  {\begin{enumerate}%
    \setlength{\itemsep}{0pt}%
    \setlength{\parskip}{0pt}}%
  {\end{enumerate}}

\newenvironment{itemize*}%
  {\begin{itemize}%
    \setlength{\itemsep}{0pt}%
    \setlength{\parskip}{0pt}}%
  {\end{itemize}}

\title{Exploiting Embodied Simulation to Detect Novel Object Classes Through Interaction}
 
\author{{\large \bf Nikhil Krishnaswamy (nkrishna@colostate.edu)} \\
  Department of Computer Science, 1873 Campus Delivery \\
  Fort Collins, CO 80523 USA
  \AND {\large \bf Sadaf Ghaffari (sadafgh@colostate.edu)} \\
  Department of Computer Science, 1873 Campus Delivery \\
  Fort Collins, CO 80523 USA
  }

\begin{document}

\maketitle

\begin{abstract}
In this paper we present a novel method for a naive agent to detect novel objects it encounters in an interaction. We train a reinforcement learning policy on a stacking task given a known object type, and then observe the results of the agent attempting to stack various other objects based on the same trained policy. By extracting embedding vectors from a convolutional neural net trained over the results of the aforementioned stacking ``play,'' we can determine the similarity of a given object to known object types, and determine if the given object is likely dissimilar enough to the known types to be considered a novel class of object. We present the results of this method on two datasets gathered using two different policies and demonstrate what information the agent needs to extract from its environment to make these novelty judgments.

\end{abstract}

\section{Introduction}

Humans are efficient at seeking out experiences that are maximally informative about their environment \cite{markant2014better,najemnik2008eye,nelson2010experience,renninger2007look,schulz2007serious}.  We explore the physical world to practice skills, test hypotheses, learn object affordances, etc. \cite{caligiore2008using,gopnik1997words,gopnik2010babies,gopnik2012scientific,gottlieb2018towards,neftci2019reinforcement,piaget1963attainment,piaget2008psychology,son2006metacognitive}. Young children, in particular, can rapidly expand their vocabulary of concepts with few or no examples, and generalize from previous to new experiences \cite{clark2006language,colung2003emergence,vlach2012fast}.

Meanwhile, artificial neural networks require large numbers of samples to train. It may take 5-8 layers of artificial neurons to approximate a single cortical neuron \cite{beniaguev2020single}. Common few-, one-, or zero-shot learning approaches in AI provide at best a rough simulacrum of human learning and generalization \cite{knudsen1994supervised,niv2009reinforcement,zador2019critique}. Recent successes in few-shot learning in end-to-end deep neural systems still require extensive pre-training and fine-tuning, often on special hardware \cite{brown2020language} or specific task formulation \cite{schick2020s}. They do not easily or organically expand to accommodate new concepts.


In this paper, we present a method to rapidly detect the introduction of a new class into an environment. We use a mixture of reinforcement learning (RL) for a stacking task in an embodied simulation environment, convolutional neural networks, and analysis of high-dimensional vector spaces to determine when the behavior of an object in interaction is inconsistent enough with the expected behavior of known object classes to be considered a likely novel class of object.  Our experiments reveal that machine learning and simulation can be leveraged for their relative strengths in this task to quickly bootstrap new models, and that making implicit information about object {\it habitats} \cite{pustejovsky2013dynamic} and {\it affordances} \cite{gibson1977theory} available to the model is critical to its performance.

\section{Related Work}

This work relates to three primary areas: object recognition and classification, embodied interaction, and reinforcement learning for simple tasks of the kind that toddlers and small children are able to solve and learn from.  Object recognition and classification is of course a well-traveled area in AI, but the AI approaches also have antecedents in the cognitive science community.  Among many others, \citeA{riesenhuber2000models} presented models of computational object recognition inspired by processes in the human visual cortex, \citeA{oliva2007role} motivated development on pre-neural network computer vision systems through examining human use of contextual cues in object recognition, and \citeA{dicarlo2007untangling} drew on both neurophysiology and computation to examine the brain mechanisms that allow for rapid object recognition under multiple circumstances.

Among approaches where the interaction between agent and environment are central, \citeA{nolfi2005category}, drawing on ``embodied cognitive science'' \cite{scheier1999embodied}, proposed a theory of category formation based on the results of interacting with the environment in simple tasks.  \citeA{bar2006functional} used simulation to classify objects based on their functional properties, but did not look at identifying when a novel class has been introduced.

Learning to stack is, of course, not a novel task in the RL community (cf. \citeA{lerer2016learning}, \citeA{li2017acquiring}, \citeA{li2020towards}, \citeA{hundt2020good}, just to name a few).  While it is useful task for demonstrating RL algorithms and AI's capability to learn representations of certain physical intuitions, the work we present here also demonstrates how an RL model for this relatively simple task, coupled with embodied simulation, can be used to drive computational implementations of certain metacognitive processes.

\section{Methodology}

Our methodology to detect novel objects can be summarized as follows:

\begin{enumerate}
    \item Train a policy to perform a task with a known object type.
    \item Attempt to use any object presented in the same task using the aforementioned trained policy.
    \item Observe differences in behaviors of the various objects and use those differences to identify if an instance of an object is sufficiently different from known objects to likely constitute a new class.
\end{enumerate}

As we are attempting to approximate certain metacognitive aspects of infant and toddler learning, we use as our task a common activity for toddlers: stacking blocks.  At slightly more than 6 months old, most infants appear able to intuit than an object will not fall if supported from the bottom on over 50\% of its lower surface \cite{baillargeon1992development,dan2000development,huettel2000effects,spelke2007core}. Therefore, an RL algorithm should be able to solve for a policy that resembles this intuition in a stacking task.  We train our stacking policy using the VoxSim simulator developed by \citeA{krishnaswamy2016voxsim}, which provides an integration with Unity ML-Agents \cite{juliani2018unity}, OpenAI Gym \cite{brockman2016openai}, and the Stable-Baselines3 reinforcement learning package \cite{raffin2019stable}.  VoxSim is based on the VoxML modeling language \cite{pustejovsky2016voxml}, which models, among other things, the rotational and reflectional symmetry of objects, which determines in part how they behave under interaction.

\subsection{Policy Training}

\begin{figure}
    \centering
    \includegraphics[height=1in]{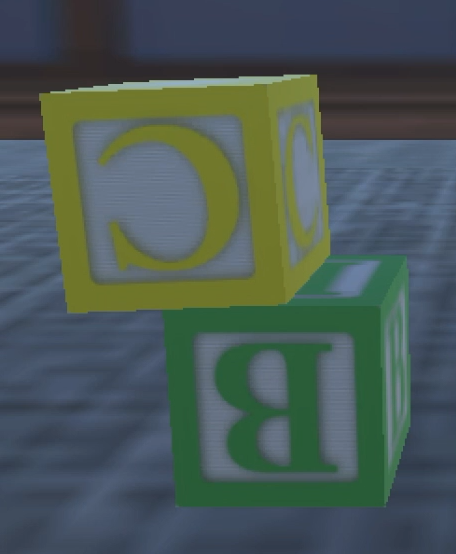}
    \includegraphics[height=1in]{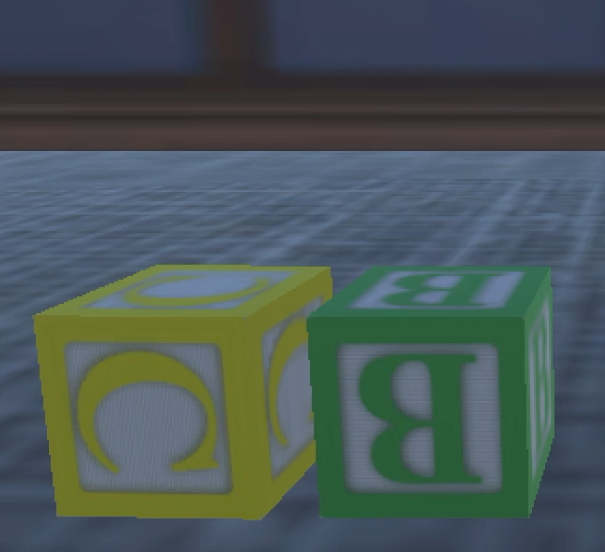}
    \includegraphics[height=1in]{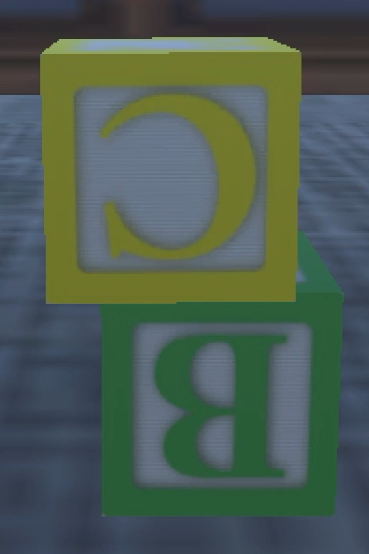}
    \vspace{-2mm}
    \caption{An unsuccessful stacking attempt (left and middle), followed by a successful one (right).}
    \vspace{-2mm}
    \label{fig:stack-training}
\end{figure}

We first train a TD3 policy \cite{fujimoto2018addressing} to stack two equally-sized cubes.  One cube is selected as the destination object and the other as the {\it theme} object (object that is moved).  The scaled action space is a 2D continuous space $[0,n]\times[0,n]$, where an arbitrary value $m_d$, where $0 \leq m_d \leq n$, represents the optimal action in dimension $d$.  The optimal action is that which places the theme object directly centered atop the destination object.  By default $m_d = \frac{1}{2}n$ (though this can be perturbed in our VoxSim agent implementation to test generalization) and we train our policy using $n = 1000$. The values of the action determine where the theme object is placed relative to the destination object, such that values close to $m_d$ will place the theme close to centered atop the destination object and values close to $0$ or $n$ will miss the surface of the destination object entirely.  The agent takes an action to place the theme block and waits to see if the two-block stack will stay up or fall down (Fig.~\ref{fig:stack-training}).

After the action is complete, we also add a small ``jitter'' to the object.  Since our agent simply moves blocks in space in the virtual environment, this simulates the small force that a real embodied agent (i.e., a toddler, or a robot) would exert upon the object when releasing it. This makes the simulation more realistic. This jitter force is applied perpendicular to the major rotational axis of the theme object if one exists.  Since cubes are symmetrical along all 3D axes, this force in training is applied in a random direction.

\begin{figure}
    \centering
    \includegraphics[width=.45\textwidth]{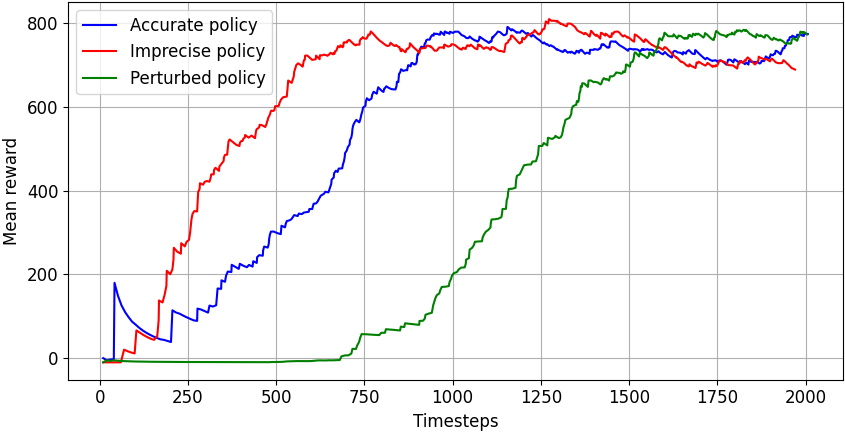}
    \vspace{-2mm}
    \caption{Episode mean reward vs. training time. In the plot where the reward starts climbing around timestep 700, the action space was perturbed so the optimal policy is far from the center.}
    \vspace{-2mm}
    \label{fig:rl-training}
\end{figure}

The state space comprises the height of the stack in number of blocks (so always an integer 1 or 2), and the 2D center of gravity of the stack (X and Z values only) relative to the center of the destination object. The agent receives a reward of -1 for missing the destination block entirely, a reward of 9 for touching the surface of the destination block but not stacking stably, and a reward of up to 1000 for stacking the two blocks successfully so they do not fall down after action completion.  The episode terminates on a successful stacking, or if the agent has tried 10 times (10 timesteps) without success. For each attempt the reward for successful stacking is decreased by 100 (i.e., 1000 for stacking successfully the first time, 900 the second time, etc.).  Fig.~\ref{fig:rl-training} shows reward plots for policies trained using this method.  The two policies that we evaluate when gathering the datasets for this paper are represented by the left two curves, each trained for 2000 timesteps. We refer to these as the {\bf accurate policy}, where the trained policy is very close to the optimal action, and the {\bf imprecise policy} where the trained policy is slightly less well-optimized and the theme cube tends to fall off somewhat more often in testing.

\subsection{Policy Evaluation and Data Gathering}

\begin{figure}
    \centering
    \includegraphics[width=.15\textwidth]{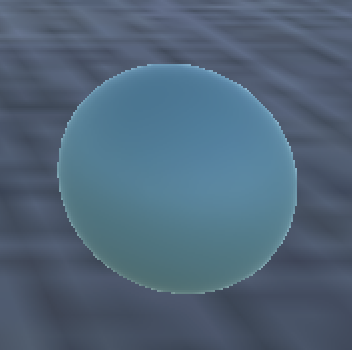}
    \includegraphics[width=.15\textwidth]{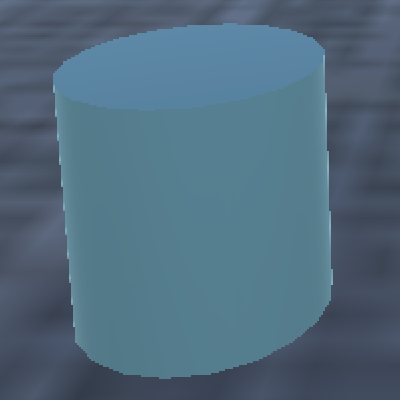}
    \includegraphics[width=.15\textwidth]{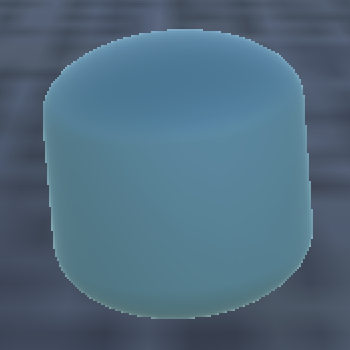}
    \vspace{-2mm}
    \caption{Sphere, cylinder, and capsule.}
    \vspace{-2mm}
    \label{fig:sphere-cyl-cap}
\end{figure}

We then evaluate the trained policies using sets of different theme objects and gather data about each evaluation.  Since the policies were trained to only stack a cube on another cube, this is tantamount to making the agent attempt to stack various objects {\it as if they are all cubes}, since policies trained for cubes are all it knows.

Besides cubes, the theme objects we evaluate with are {\it sphere}, {\it cylinder}, and {\it capsule} (Fig.~\ref{fig:sphere-cyl-cap}), which are all the same size as cube but have different geometric properties.

The sphere will almost never be able to be stacked since it will roll off.  The cylinder can often be stacked if placed in the right location upright, but will usually (not always) roll off the bottom cube if placed horizontally.  Thus, it shares some behavior with both a cube and a sphere.  The capsule will almost always fall off (like sphere) if placed vertically but might occasionally remained stacked if placed horizontally (like cylinder).  We also evaluate using a theme object that is a {\it small cube}, one quarter the volume of the destination cube. With each of these objects we should see distinct behavior in aggregate when attempting to stack them.

The rotational axis used in computing the direction of the post-action jitter is encoded in the VoxML semantics of the theme object, as shown in Fig.~\ref{fig:cylinder-typing}, therefore if the theme object is a cylinder, the post-action release jitter is applied perpendicular to its local Y-axis.

\begin{figure}[h!]
\vspace{-2mm}
{\tiny
\def\baselinestretch{1.1}
$\avmplus{\att{cylinder}\\
	\attval{type}{\avmplus{
		\attvaltyp{head}{cylindroid}\\
		\attvaltyp{components}{nil}\\
		\attvaltyp{rotatSym}{$\{Y\}$}\\
		\attvaltyp{reflSym}{$\{XY,YZ\}$}
	}}                                
}$
\def\baselinestretch{1.9}
}
\vspace{-2mm}
\caption{VoxML typing structure for a cylinder, showing axes and planes of rotational and reflectional symmetry.}
\label{fig:cylinder-typing}
\vspace{-2mm}
\end{figure}

\begin{figure*}
    \centering
    \includegraphics[width=.195\textwidth]{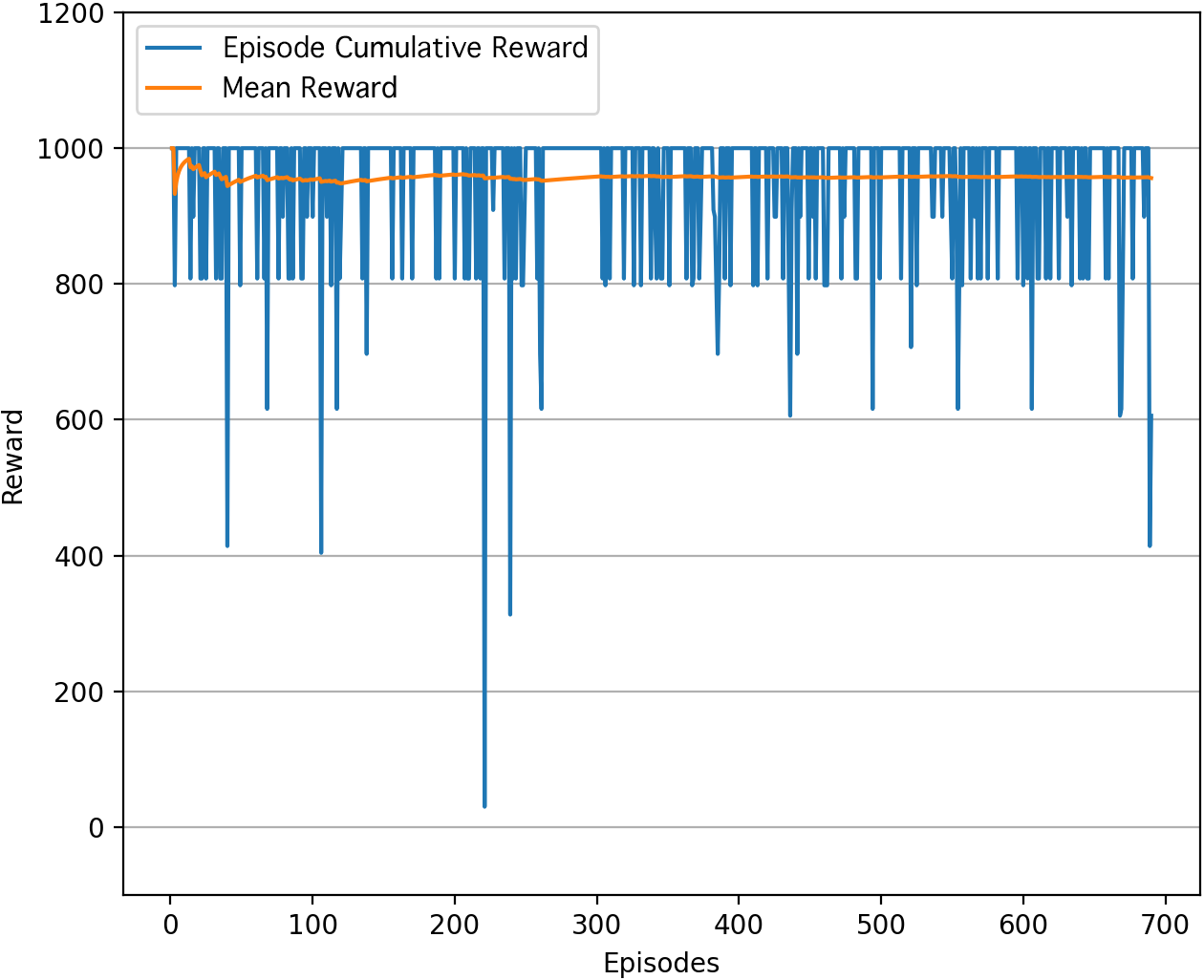}
    \includegraphics[width=.195\textwidth]{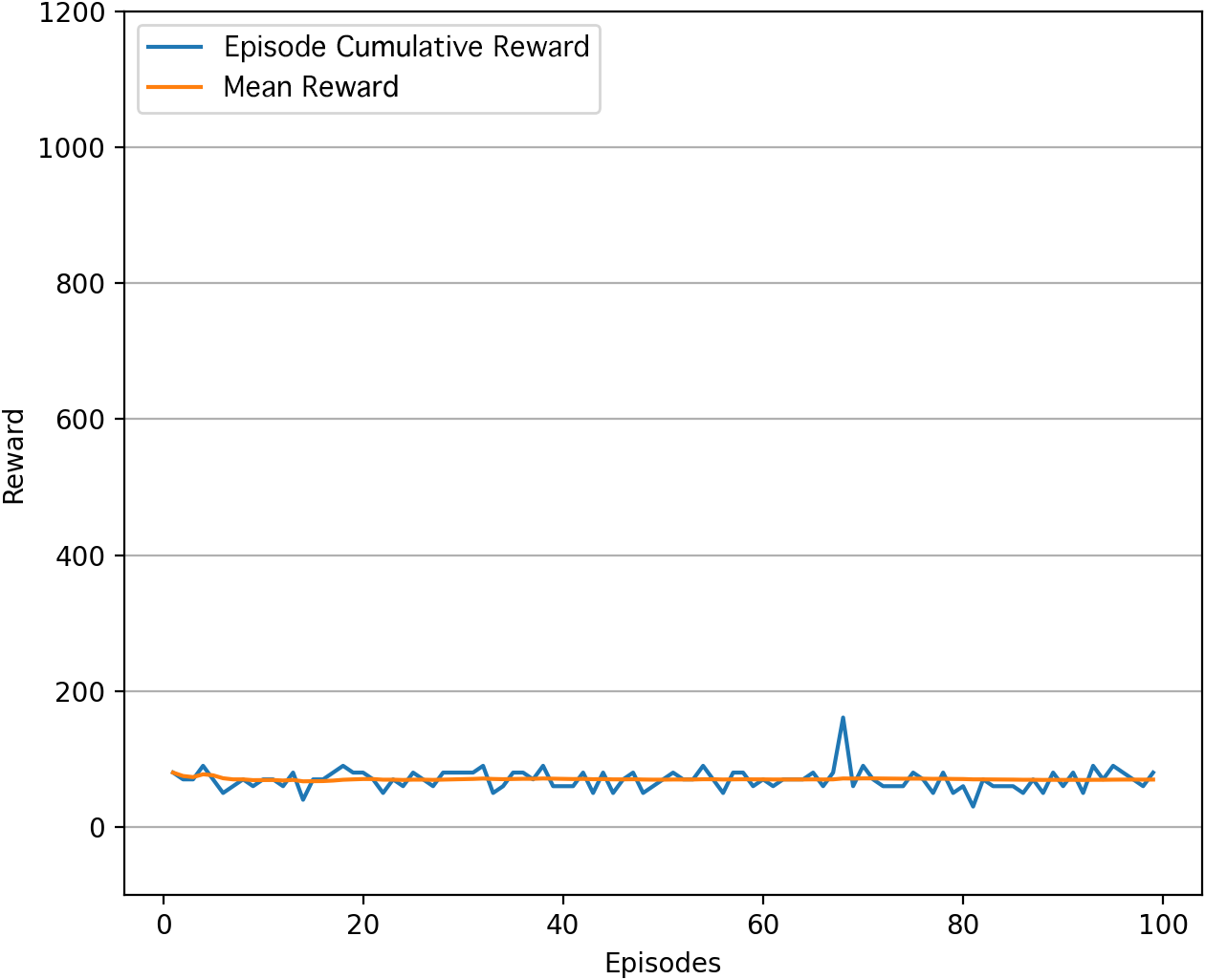}
    \includegraphics[width=.195\textwidth]{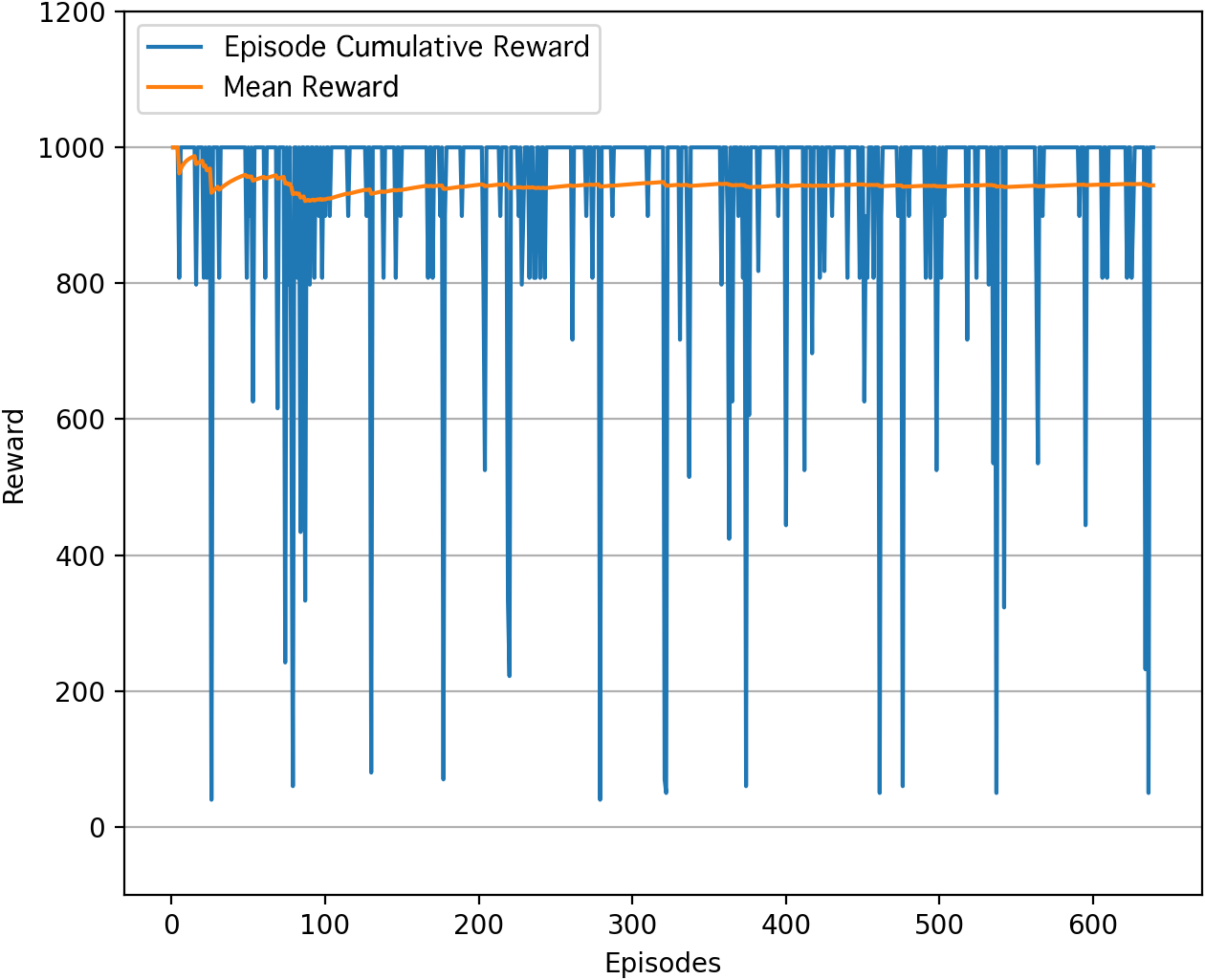}
    \includegraphics[width=.195\textwidth]{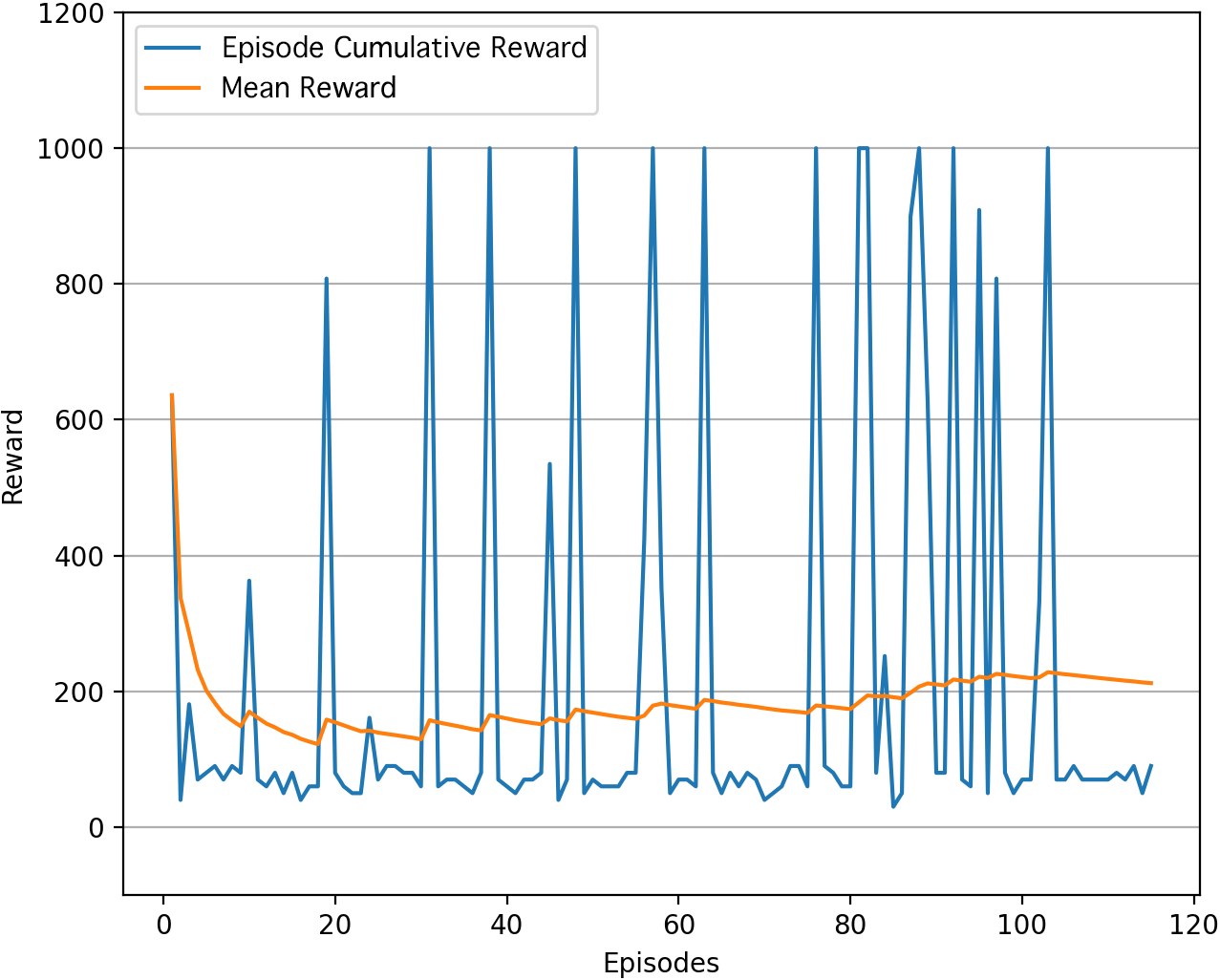}
    \includegraphics[width=.195\textwidth]{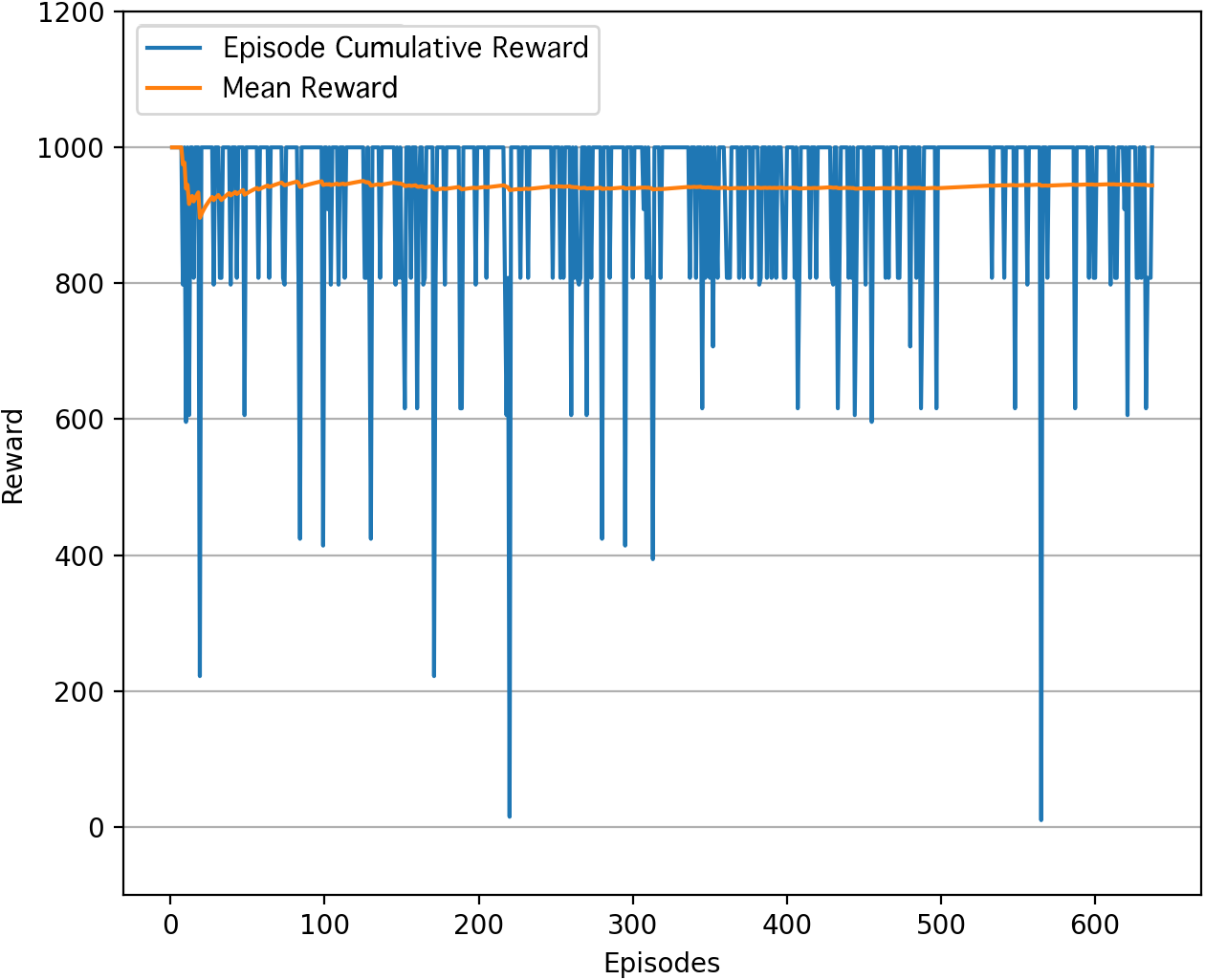}
    \vspace{-2mm}
    \caption{Reward plots for stacking (left to right) cube, sphere, cylinder, capsule, and small cube on a cube. Episodes are on the X-axis and rewards are along the Y-axis.}
    \vspace{-2mm}
    \label{fig:reward-plots}
\end{figure*}

We evaluate each policy for 1000 timesteps with each object.  Fig.~\ref{fig:reward-plots} shows the evaluation reward plots (the blue line is the reward after each episode and the orange line is the mean cumulative reward), and show that the cube is the easiest object to stack, followed by cylinder, capsule, and finally sphere.  We can also see that there is not much difference between the stackability of a big cube and a small cube, as expected. As during training, the agent gets 10 attempts to stack per episode, so stackable objects like cube and cylinder can complete more episodes in 1000 evaluation timesteps.

During evaluation, we also gather information about each stacking attempt from the VoxSim virtual environment.  At each timestep we store the type of the theme object, its rotation in radians at episode start, radians between the world upright axis and the object upright (+Y) axis, the numerical action executed, the object rotation and offset from world upright after the action, the vector of the VoxML-derived post-action jitter applied, the state observation after action completion, the reward for the attempt, the cumulative total reward over the episode, and the cumulative mean reward over the episode.  We gathered two datasets, one each using using the accurate and imprecise policy.

\subsection{Object Similarity Analysis}

Canonical correlation analysis (CCA) is concerned with finding basis vectors for two sets of multidimensional variables in an unsupervised manner, such that the correlation coefficient between the projections of the variables onto the basis vectors is maximized \cite{hotelling1992relations}. To expose the kinds of differences we want a model to find when detecting novel classes, we use CCA to find correlations between the parameters describing each object's behavior in the stacking task. It is these properties of the object and action that allow us to distinguish not just that the objects behave differently, but {\it how}. We applied CCA to the data describing each pair of objects.  Table~\ref{tab:cca} shows the correlation coefficients between each pair, averaged across the two datasets.

\begin{table}
    \centering
    \begin{tabular}{|l|l|l|l|l|l|}
    \hline
        \vspace{-1mm}
        & & & & & {\sc Sm.} \\
        & {\sc Cube} & {\sc Sph.} & {\sc Cyl.} & {\sc Cap.} & {\sc cube} \\
        \hline
        {\sc Cube} & 1 & 0.396 & {\bf 0.958} & 0.686 & {\bf 0.808} \\
        \hline
        {\sc Sphere} & 0.399 & 1 & 0.366 & {\bf 0.832} & 0.376 \\
        \hline
        {\sc Cylinder} & {\bf 0.974} & 0.366 & 1 & 0.692 & 0.528 \\
        \hline
        {\sc Capsule} & 0.688 & {\bf 0.832} & 0.692 & 1 & 0.511 \\
        \hline
        {\sc Sm. cube} & {\bf 0.808} & 0.376 & 0.527 & 0.511 & 1 \\
    \hline
    \end{tabular}
    \vspace{-2mm}
    \caption{CCA between object pairs (averaged across datasets). Greatest correlation coefficient in each row and column is bolded (excluding diagonal)}
    \vspace{-2mm}
    \label{tab:cca}
\end{table}

CCA exposes some of the expected correlations (or lack thereof) between object types.  Cubes, which usually stack successfully, and spheres, which almost never do, have a low correlation coefficient, whereas cylinders, which also frequently stack successfully, have a high correlation coefficient with cubes.  Spheres and capsules, which also have similar behavior (largely unstackable), have a high correlation coefficient.

Therefore we can see that CCA can expose similarities between object classes, but also might conflate or split object classes incorrectly: if a small cube is a member of the same class as big cube, cylinder must be too, because cylinder has a greater correlation coefficient with big cube than small cube does. We need a model that considers objects at a finer-grained level than raw data matrices, as many individual parameters (e.g. numerical action) are by design mostly consistent across all classes. CCA correlates the linear relationships between multidimensional variables, and is an imperfect discriminator in this task, however, it consistently shows the expected dissimilarity of cubes and spheres, which provides a starting point for the novel class detection task.

\subsection{Novel Class Detection}

We want to be able to give an algorithm a model of a subset of these classes (e.g., cube and sphere) and have it identify that a new type of object (e.g., cylinder or capsule) is different from any of the known classes based on the way it behaves when interacted with (i.e., stacked).  We also want to be able to identify that new samples of a known class (e.g., small cubes), are not new classes of objects, but additional instances of a known class.  Here we do not consider size as a distinguishing feature in the model, only object behavior when stacked.

Novel class detection follows the following procedure:

\begin{enumerate}
    \item Identify which known class an object is most similar to.
    \item Determine if the new object is different enough from the most similar known class to be considered likely novel.
\end{enumerate}

Choosing one out of a set of known classes is an obvious task for a forced-choice classifier. We start by training a classifier on two known classes: cube and sphere, the two most dissimilar objects according to CCA.  We use a 1D convolutional neural net for this task, written in PyTorch.  We group inputs by episodes and to maintain a balanced sample, use only the first 90 testing episodes, reserving a further 10 as a development set for testing the classifier, and the remainder of the data for detecting novel classes.  Since episodes can be variable length (depending on how successful the policy was at stacking the object in question), we pad out the length of each input to 10 timesteps, copying the last sample out to the padding length.  Therefore an episode where the policy stacked the object successfully on the first try will consist of 10 identical timestep representations, while an episode where the agent tried and failed the stack the object 10 times will have 10 different timestep representations.

\begin{figure}
    \centering
    \includegraphics[width=.45\textwidth]{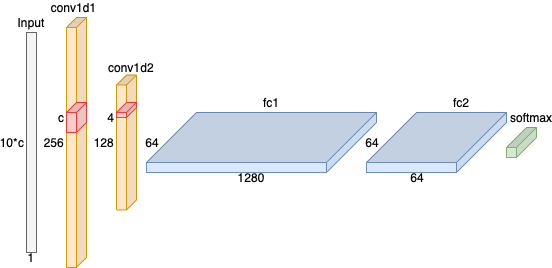}
    \vspace{-2mm}
    \caption{1D CNN object classifier diagram.}
    \vspace{-2mm}
    \label{fig:cnn}
\end{figure}

The classifier consists of two convolutional layers (256 and 128 hidden units respectively). The filter size in the first layer is $c$, a variable equal to the number of parameters saved at each evaluation timestep during data gathering ($c = 19$ here) and a stride length of 8, and the second layer uses a filter size of 4 and a stride length of 2.  This allows the convolutions to generate feature maps in the hidden layers that are approximately equal to the size of a single timestep sample, and convolving over this approximates observing each timestep of the episode in turn.  The convolutional layers are followed by two 64-unit fully-connected layers and a softmax layer.  All layers use ReLU activation.  Fig.~\ref{fig:cnn} shows a network diagram.  We train for 500 epochs using the Adam optimizer \cite{kingma2014adam}, a batch size of 100 ($= 10$ episodes) and a learning rate of 0.001.

Since the differences between cubes and sphere in their stacking behavior are so evident, this simple two-class classifier can routinely achieve 100\% evaluation accuracy on the dev-test set.

We then take batches drawn from classes unknown to the model, e.g., cylinder and capsule, and from additional instances of known classes, e.g., cubes, spheres, and small cubes.  The classifier, trained over only two classes, will classify even the non-sphere or cube samples as sphere or cube.  Most commonly, cylinder is classified as cube and capsule is classified as sphere, because these objects' stacking behavior is similar.  Small cube is (correctly) classified as a cube.

We then go into the final fully-connected layer of the network and pull out the 64-dimensional {\it embedding vectors} for each sample in the testing batch, and for each sample of the most similar known class.  Regardless of the class the classifier predicts, the embedding vectors of the new samples can be compared to embedding vectors of known instances of that predicted class to determine if this new batch is similar enough to truly be the same as the known class or not.

We compute $\vec{\mu_S}$ and $\vec{\sigma_S}$, the mean embedding vector of the known class and the standard deviation of the known class vectors, respectively, as well as $\vec{\mu_N}$, the mean embedding vector of the new batch.

Then, assuming that if all samples, new or known, were in fact members of the same class, there would still be some outliers, we find individual outliers in the new batch samples and in the known class samples by dividing the cosine distance between $\vec{\mu_S}$ and the sample $\vec{v}$ in question by the cosine distance between $\vec{\mu_S}$ and $\vec{\mu_S}+\vec{\sigma_S}$. Let $\rho_{\vec{v}} = \frac{\text{cos}(\vec{\mu_S},\vec{v})}{\text{cos}(\vec{\mu_S},\vec{\mu_S}+\vec{\sigma_S})}$, and if $\rho_{\vec{v}} > 1$, the sample $\vec{v}$ is considered to be an outlier $\vec{o} \in O$, where $\rho_{\vec{o}} = \frac{\text{cos}(\vec{\mu_S},\vec{o})}{\text{cos}(\vec{\mu_S},\vec{\mu_S}+\vec{\sigma_S})}$.  Let $O_S$ be the set of outliers among the samples of the known class $S$ and let $O_N$ be the set of vectors in the new batch $N$, where $\rho_{\vec{o_N}} > 1$.  Outlying samples may still belong to the known class (e.g., sometimes a cube simply fails to stack properly due to bad placement, not its properties, but nonetheless appears to be very different from other cubes in terms of its behavior), so we perform Z-score filtering on the computed outliers, using a Z threshold of 3, and $\mu_\rho$ and $\sigma_\rho$, the mean and standard deviation, respectively, of the previous computations over the outlier vectors.  If $\frac{(\rho_{\vec{o}}-\mu_\rho)}{\sigma_\rho} \geq 3$, $\vec{o}$ is removed from the set of outlier embeddings.  For all outliers $\vec{o_N} \in O_N$ that were derived from new batch samples and all outliers $\vec{o_S} \in O_S$ derived from known class samples, we sum $\rho_{\vec{o_N}}$ for all $\vec{o_N} \in O_N$ and divide by the sum of $\rho_{\vec{o_S}}$ for all $\vec{o_S} \in O_S$.  This produces an ``outlier ratio'':

\begin{center}
$OR = \frac{\sum_{\vec{o_N} \in O_N} \rho_{\vec{o_N}}}{\sum_{\vec{o_S} \in O_S} \rho_{\vec{o_S}}}$    
\end{center}

Finally, we multiply the outlier ratio by $\text{cos}(\vec{\mu_S},\vec{\mu_N})$ (the cosine distance between the mean of the known samples and the mean of the new batch), divide that by $\text{cos}(\vec{\mu_S},\vec{\mu_S}+\vec{\sigma_S})$ (the cosine distance between the mean of the known samples and the mean of the known samples plus their standard deviation), and multiply that by the denominator of the outlier ratio.  This approximates how many times more dissimilar a given batch is from the mean of the known class than a random sample that falls within the vector substance spanned by the known class samples would be.  Given that a new sample of a known class may not fall exactly within the vector subspace spanned by the samples in the data, we want this dissimilarity threshold to be greater than 1, acknowledging that the subspace defining a class may expand as new samples belonging to that class are encountered.  Therefore we define a dissimilarity threshold $T$, and if

\begin{center}
$\frac{OR \times \text{cos}(\vec{\mu_S},\vec{\mu_N})}{\text{cos}(\vec{\mu_S},\vec{\mu_S}+\vec{\sigma_S}) \times \sum_{\vec{o_S} \in O_S} \rho_{\vec{o_S}}} > T$    
\end{center}
we say that the batch of new samples likely belongs to a class that is not one of the known classes.

\section{Results}

\begin{figure}[h!]
    \centering
    \includegraphics[width=.45\textwidth]{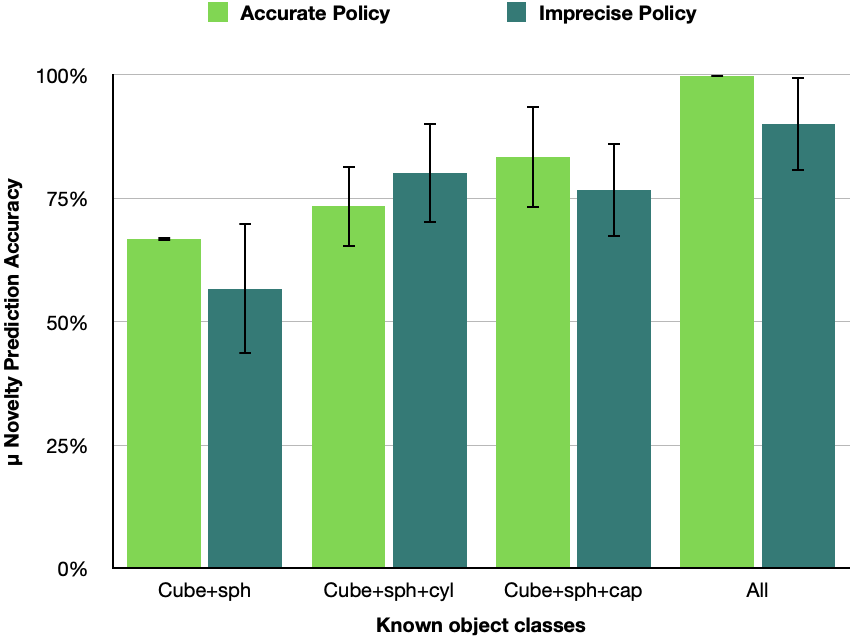}
    \includegraphics[width=.45\textwidth]{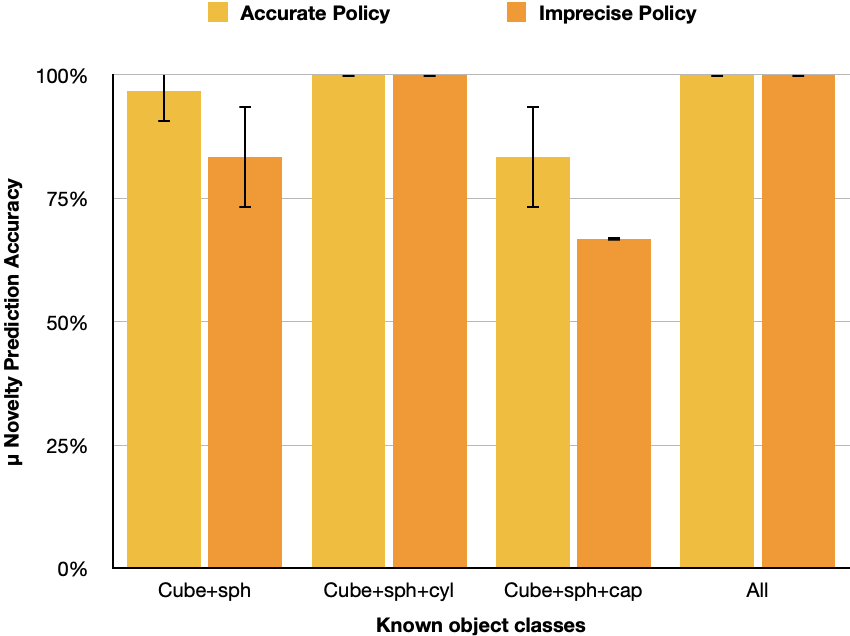}
    \vspace{-2mm}
    \caption{Novel class identification accuracy under each condition. Top chart shows results without the VoxML-derived jitter information, and bottom chart shows results with it.}
    \vspace{-2mm}
    \label{fig:novel-acc}
\end{figure}

Here we present results of the method detailed above.  We implemented tests where the classifier model was trained to classify different sets of known classes (cube and sphere, cube and sphere and cylinder, cube and sphere and capsule, and all four).  We also conducted an experiment where we trained the 1D CNN without the VoxML-derived jitter force information, which implicitly encodes the axis of symmetry of the theme object, to compare what information this adds to the model.  We conducted 10 experiments under each condition with each dataset, using a dissimilarity threshold value of $T = 25$.  Correct results were identifying cylinder and capsule as novel classes where they were not already known, and not identifying small cube as a novel class.  Fig.~\ref{fig:novel-acc} shows the aggregate results with confidence intervals.

We can achieve high accuracy in all cases, correctly identifying the novelty of cylinders and capsules where appropriate, and identifying batches of small cubes as instances of the known cube class, simply based on the way they behave in the stacking task.  The imprecise policy data is somewhat more challenging, because even stable objects like cubes fall off the bottom cube more often due to bad placement.

\section{Discussion}

First, it is clear that using the VoxML-derived information in the object classifier provides a strong boost to the novel class identification performance, often up to 25\%.  This is largely because without this information, the model cannot infer which axis the theme object moved along when the jitter force was applied, and therefore cylinder embedding vectors end up nearly identical to cube embeddings in most respects.  Therefore it seems that this use of VoxML is encoding information useful for common-sense reasoning a la \citeA{hobbs1984sublanguage} into the model.

Second, it is clear that when cylinder is a known class to the model, it is much easier to identify capsule as a novel class than it is to identify cylinder as a novel class when capsule is given.  This suggests that the order of class acquisition is important.  When the VoxML inputs are included in classifier training, our method can identify capsule as a novel class 100\% of the time in our experiments, but when capsule is known first, the ability to acquire cylinder is impeded to 75\% accuracy or less.  We hypothesize that this is because cylinder-to-cube is a much more fine-grained distinction than capsule-to-cube or capsule-to-sphere.  Because capsule is more markedly different in its behavior compared to cube or sphere, the capsule vectors take up a lot more space in the overall embedding space, making it difficult for cylinder embeddings to be distinguished from other classes (usually cubes).

\begin{figure}
    \centering
    \includegraphics[width=.2\textwidth]{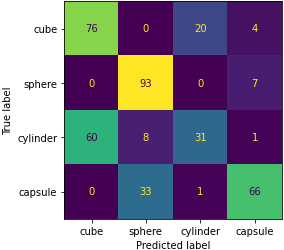}
    \includegraphics[width=.2\textwidth]{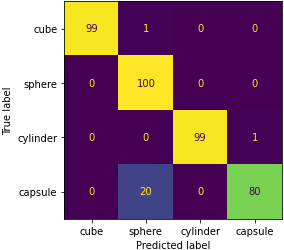}
    \includegraphics[width=.2\textwidth]{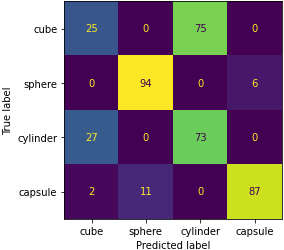}
    \includegraphics[width=.2\textwidth]{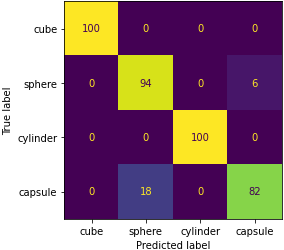}
    \vspace{-2mm}
    \caption{Aggregated CNN classifier outputs over the dev-test set. Top: accurate policy evaluation; Bottom: imprecise policy evaluation; Left: without VoxML-derived inputs; Right: with VoxML-derived inputs.}
    \vspace{-2mm}
    \label{fig:classifier-cms}
\end{figure}

Finally, even without the VoxML-derived information available to the model, the novel class detection method can do a good job of determining that the small cube is not a novel class, as witnessed by the right two bars in Fig.~\ref{fig:novel-acc}.  However, a deeper look into the classifier outputs show that even though the small cubes are being correctly labeled as ``not novel,'' the most similar class they are subsumed into is often not {\it cube} but {\it cylinder}.  Therefore it is clear that the VoxML-derived inputs are critical to correctly classifying the behavioral distinctions between objects like cubes and cylinders in the first place, in order to correctly assess the novelty of these classes.  Fig.~\ref{fig:classifier-cms} shows the CNN classifier outputs over the 10 episode dev-test set, aggregated over all 10 novel concept detection experiments we conducted.  The two confusion matrices on the left show classification results without the VoxML-derived inputs.  The two on the right show results with those inputs.  The top two are from the accurate policy evaluation, and the bottom two are from the imprecise policy evaluation.  Without the VoxML-derived inputs, we see frequent confusion between sphere and capsule, and more so between cube and cylinder, so clearly these inputs when extracted from the environment are important for the success of this task.

\section{Conclusion and Future Work}

We have presented a method for a naive agent to detect the introduction of novel classes of objects into its environment.  We use a combination of reinforcement learning, neural networks, and statistical methods to rapidly identify when an object is likely to belong to a novel class based on how it behaves during an interaction.  We have presented results that demonstrate how we can do this with high accuracy, and a low false-positive and false-negative rate.  We have also presented evidence of what kind of information is important to capture for the success of this task, and how the order of object class identification is potentially important.

As mentioned, we intend this method to approximate certain aspects of infant and toddler learning.  One critical difference between human learning and AI learning is that most modern AI techniques require very large volumes of data, long training times, or specialized (often expensive) hardware.  The models and processes we have developed here are small, lightweight, and can be trained using only about 1000 individual timestep samples per known class.  All models and calculations presented here can be trained or performed on a laptop using the CPU, though GPU training provides a speed benefit, and a potential step toward a computational ``fast-mapping'' style of concept acquisition, especially as more classes are acquired.\footnote{Full code and environment will be released upon deanonymization.}

In future work, we would like to pursue a curriculum learning approach, where the convolutional layer weights of the classifier are further trained to optimize for classifying a novel class once it is identified, and then testing the capacity for the new model to generate embeddings that can be used to detect subsequent novel classes.

Currently, the dissimilarity threshold we use is held constant.  We hypothesize that the best dissimilarity threshold is likely to change as more classes are identified, as sections of the vector space become associated with subspaces defining certain classes.  Therefore we will pursue methods for calibrating the best dissimilarity threshold for the data at hand.

We use a convolutional network approach, which involves padding the data.  We would also like to investigate a recurrent approach that could consume variable-sized inputs, and investigate its effect on the nature of the extracted embeddings.  Other future directions include: different statistical and geometric techniques for assessing similarity, the incorporation of inputs that correlate with size, allowing us to potentially identify small cubes as distinct by virtue of features other than stacking behavior; and starting with a model that has knowledge of only one class, rather than two.



\bibliographystyle{apacite}

\setlength{\bibleftmargin}{.125in}
\setlength{\bibindent}{-\bibleftmargin}

\bibliography{cogsci_full_paper_template}

\end{document}